\documentclass{article}
\usepackage{spconf,amsmath,graphicx,multirow,amssymb}
\usepackage{epsfig}
\usepackage{graphicx}
\usepackage{multirow}
\usepackage{amsmath}
\usepackage{amssymb}
\usepackage{subcaption}
\usepackage{comment}
\usepackage{adjustbox}
\usepackage{bbm}
\usepackage{color}
\usepackage[ruled]{algorithm2e}

\title{DIRECT: Deep DIscRiminative Embedding for ClusTering of LIGO Data}
\name{S.~Bahaadini$^\star$, V.~Noroozi$^{\star\star}$, N.~Rohani$^\star$, S.~Coughlin$^\dagger$, M.~Zevin$^\dagger$, and A.~Katsaggelos$^\star$}
\address{$\star$  Electrical Engineering and Computer Science, Northwestern University, Evanston, IL, USA\\
$\star\star$  Department of Computer Science, University of Illinois at Chicago, IL, USA,\\
$\dagger$  Center for Interdisciplinary Exploration and Research in Astrophysics (CIERA) and \\Dept. of Physics and Astronomy, Northwestern University, Evanston, IL, USA}

\begin{document}
%
\maketitle
\begin{abstract}
In this paper, benefiting from the strong ability of deep neural network in estimating non-linear functions, we propose a discriminative embedding function to be used as a feature extractor for clustering tasks. The trained embedding function transfers knowledge from the domain of a labeled set of morphologically-distinct images, known as \textit{classes}, to a new domain within which new classes can potentially be isolated and identified. Our target application in this paper is the \textit{Gravity Spy} Project, which is an effort to characterize transient, non-Gaussian noise present in data from the Advanced Laser Interferometer Gravitational-wave Observatory, or LIGO. Accumulating large, labeled sets of noise features and identifying of new classes of noise lead to a better understanding of their origin, which makes their removal from the data and/or detectors possible.
\end{abstract}
\section{Introduction}
\label{sec:intro}
The advanced Laser Interferometer Gravitational-wave Observatory (LIGO, \cite{aLIGO}) recently made the first direct observations of gravitational waves emanating from the final orbits and merger of binary compact object systems \cite{GW150914,GW151226,O1BBH,abbott2017gw170817}. These observations require sensitivity to fractional changes of distance on the order of $10^{-21}$. Though all sensitive components of LIGO are exquisitely isolated from non-gravitational-wave disturbances, the extreme sensitivity of LIGO still makes it susceptible to disturbances that cause noise in the detectors and can afflict searches for gravitational waves. Transient, non-Gaussian noise sources known colloquially as \textit{glitches} occur at a significant rate, come in many morphologies, and can mask or mimic gravitational-wave signals. 
A comprehensive classification and characterization of these noise features is needed to identify their origin, construct vetoes to eliminate them from the data, and/or to remove their root cause from the instrument itself.

The \textit{Gravity Spy} project \cite{zevin2017gravity} is designed to classify these glitches into morphological categories by combining the strengths of machine learning algorithms and crowdsourcing.
The dataset of glitches~\cite{bahaadini2017deep} are represented as spectrogram images in time-frequency-energy space, where $22$ morphologically-distinct classes are currently accounted for \cite{zevin2017gravity}.  
In~\cite{bahaadini2017deep,zevin2017gravity,datasetpaper,daniel}, this multi-class classification problem has been tackled by the application of deep learning algorithms. In~\cite{daniel}, some initial efforts toward glitch clustering are presented.
%

Because of variable environmental conditions at the sites and changes in the sensitivity and design of the LIGO detectors over time, glitch classes are not static, and new morphological classes regularly appear in the data. Therefore, the identification of new glitch classes is a route worthy of investigation. 
By considering the morphological characteristics of glitches,
new classes can be defined and integrated into the Gravity Spy project~\cite{zevin2017gravity} which will bolster the number of labeled glitches in these new morphological classes. Updating glitch classes in such a manner will help us follow changes in the noise present in LIGO data and allow for their suppression or removal. 

In this paper, we present our model for clustering the glitches that are identified through the Gravity Spy framework as not belonging to the set of known glitch classes. 
Our suggested algorithm transfers knowledge from the domain of \textit{known} glitch classes to the domain of \textit{unknown} glitch classes. To this end, a deep neural network model is trained with the samples from known glitch classes. This neural network learns the parameters of a nonlinear embedding function that works as a feature extractor to give us a discriminative feature space where samples from the same class are close to each other while samples from different classes are far from each other. The embedding function projects samples to a discriminative space that allows the clustering algorithm to work more effectively. This enables the clustering algorithm to find potential new glitch classes over the space of unknown glitch samples.
\vspace{-2mm}
\begin{figure}[h!]
\centering
\includegraphics[width=0.25 \textwidth]{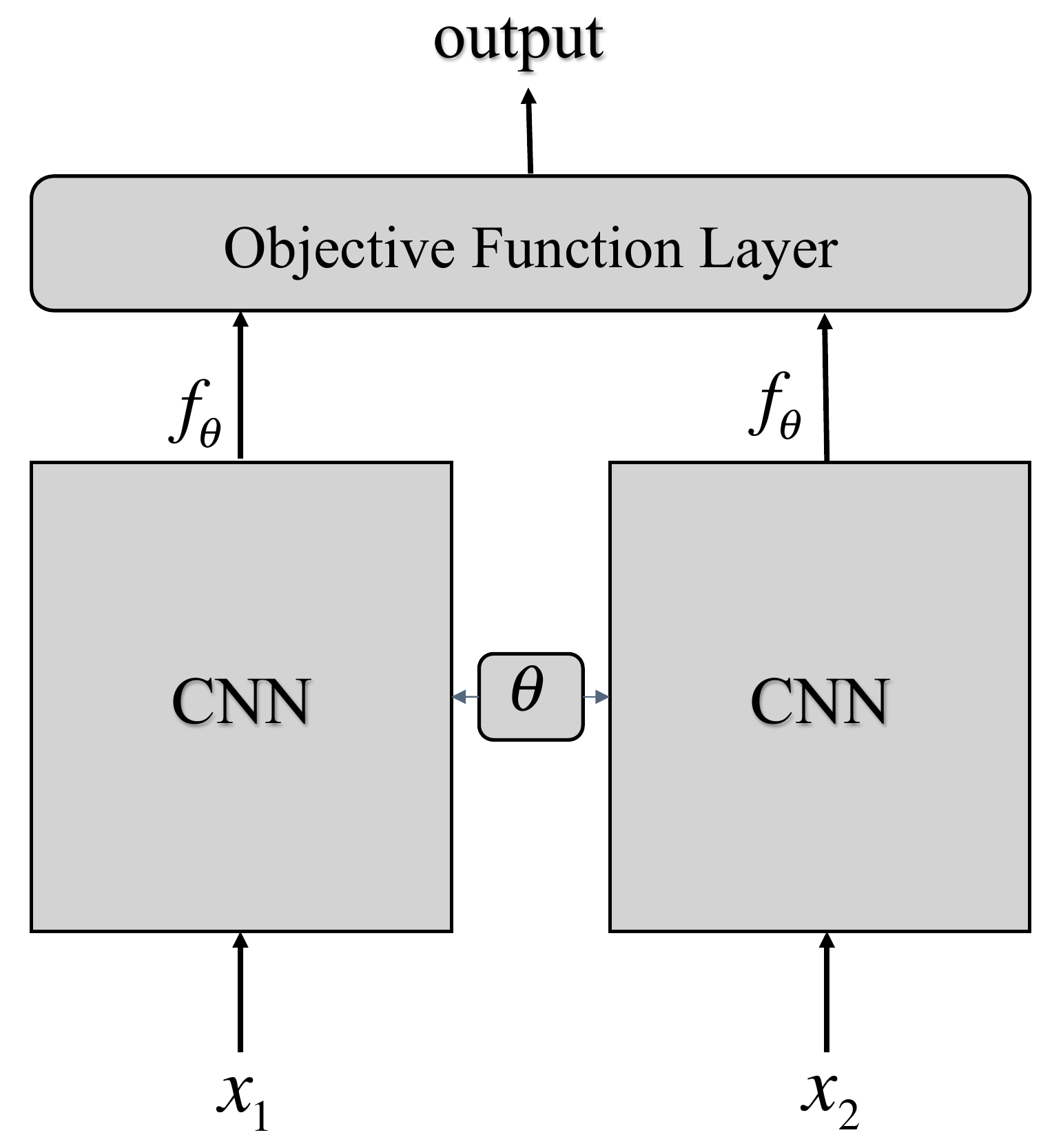}
\caption{The schematic representation of the deep neural network that is trained to yield a discriminative feature space. CNN is the convolutional neural network.}
\label{fig:model}
\end{figure}
\section{Proposed Framework}
\label{sec:alg}
The clustering task is an unsupervised machine learning algorithm~\cite{watt2016machine}. In this study, we transfer and inject knowledge to the clustering algorithm using deep neural networks. Our algorithm, which is called Deep DIscRiminative Embedding for Clustering of LIGO Data (DIRECT), uses a labeled set of glitch classes as the source domain and a pool of unlabeled glitch samples as the target domain, which may or may not belong to the glitch classes accounted for in Gravity Spy. This type of task is also referred to as domain adaptation~\cite{goodfellow2016deep}.

We define a nonlinear embedding function $f_{\theta}$, which is used to give a new discriminative representation for glitch data. A deep neural network model is implemented to learn $f_{\theta}$. The schematic representation of this model is shown in Fig.~\ref{fig:model}. The model is trained with the pairs of samples selected from the set of known glitch classes. It is trained such that in this new feature space, samples from the same class are located close to each other while samples from different classes are far from each other. This is a desirable property which is a called \textit{discriminative feature space}. We hypothesize that this discriminative space, though trained with 
samples that may be quite different from the unlabeled samples, will lead to the improved clustering of the unlabeled testing samples.
We test this hypothesis in Section~\ref{sec:exp}, and the objective function which determines this discriminative space is discussed in the following section.

\subsection{Objective Function}
\label{sec:obj}
The objective function $\mathcal{L}$ of the model is defined as
\begin{multline}
\label{eq:obj}
\mathcal{L} = \sum_{i=1}^{N}l(y^i,x_1^i,x_2^i) = \sum_{i=1}^{N} y^i \times {dist(f_{\theta}(x_1^i) - f_{\theta}(x_2^i))} 
\\+ (1-y^i)\times \textrm{max}(0, m-{dist(f_{\theta}(x_1^i) - f_{\theta}(x_2^i))})
\end{multline}

\noindent
where $N$ is the number of training pairs made from known classes, $x_1^i$ and $x_2^i$ are the first and second items of the $i^\textrm{th}$ pair, $y^i$ is the binary label of the $i^\textrm{th}$ pair which is one when the two items of the pair belong to the same class and zero when they belong to different classes, $f_{\theta}(.)$ is the nonlinear function modeled by a convolutional neural network in Fig.~\ref{fig:model}, $dist$ is a distance function (such as Euclidean or Cosine), and $m$ is the margin that is used to bound the distance between the items of pairs from different classes. This objective function was originally proposed in~\cite{siamese} for signature verification and its semi-supervised version has been proposed in~\cite{noroozi2017seven}.

\subsection{DIRECT}
\label{sec:direct}
As DIRECT contains a deep neural network, we first train the network with a training set consisting of pairs of known samples.
Given $|X|$ labeled glitch samples, we can make $\binom{|X|}{2}$ pairs. Our labeled set of $\sim$10,000 images thus leads to almost $50$ million pairs. To limit computational costs, we consider a smaller subset of pairs for training. In order to span the whole space of possible pairs better, for each epoch in the algorithm randomly choose a new set of pairs.
There exits many optimization techniques The RMSprop~\cite{rmsprop} optimizer is used for optimizing Eq.\ref{eq:obj} chosen from \cite{zeiler2012adadelta,rmsprop,opt,samira2017,vahid1}.

Through training of the deep neural network, we learn the parameters of $f_{\theta}(.)$ which is then used to project the unknown samples to the discriminative feature space. Specifically, we calculate $z = f_{\theta}(x)$, where $x\in {\rm I\!R}^{d_1 \times d_2 \times 3}$ (for which $d_1$ and $d_2$ are the first and second dimension of glitch images, respectively, and $3$ is the RGB channel dimension) and $z\in {\rm I\!R}^{k}$ (for which $k$ is the size of the projected space). Thus $f_{\theta}$ alters the dimensionality of the feature space as: ${\rm I\!R}^{d_1\times d_2 \times 3} \rightarrow {\rm I\!R}^{k}$. Then, a clustering algorithm such as k-means is employed on the new feature space. 
The general steps of the suggested method is summarized in Algorithm~\ref{alg:summary}:
\begin{algorithm}[h!]
 \caption{Summary of DIRECT.}
 \label{alg:summary}
\SetAlgoLined {
\KwIn{Training set: ${{D}} = \big\{ (x_1^i,x_2^i) \big \}_{i=1}^{N}$, \\$\quad\qquad$Label set ${{Y}} = \{y^i\}_{i=1}^{N}$, 
\\$\quad\qquad$Testing set ${U} = \big\{x^m \big \}_{m=1}^{M}$, 
}
\vspace{+1mm}
\KwOut{Embedding function $f_{\theta}:{\rm I\!R}^{d_1\times d_2 \times 3} \rightarrow {\rm I\!R}^{k}$,
\\$\quad\qquad$ Clusters label set ${L} = \big\{ c^m \big \}_{m=1}^{M}$}
\vspace{+1.5mm}
\textbf{Step1:} Training deep neural network (shown in Fig.~\ref{fig:model}) with objective function Eq.~\ref{eq:obj} using RMSprop optimization method to estimate $f_{\theta}(.)$

\vspace{+1.5mm}
\textbf{Step2:} employing $f_{\theta}$ on all $x^m \in U$ as $z^m = f_{\theta}(x^m)$

\vspace{+1.5mm}
\textbf{Step3:} performing k-means algorithm on ${{Z}= \big\{z^m\big \}_{m=1}^{M}}$ and returning the corresponding cluster labels $L$

\KwRet{$L$}}
\end{algorithm}
\vspace{-6mm}
\section{Experiment}
\label{sec:exp}
\subsection{Evaluation Measures}
We use two metrics to evaluate the performance of the clustering algorithm. 

The first, known as the Normalized Mutual information (NMI) score, is a metric quantifying the similarity between predicted clusters versus true clusters. The NMI value lies in the range of $0$ when there is no mutual information between two cluster assignments to $1$ when there is perfect correlation between two sets.
NMI is defined as $NMI = \frac{I(\rm Y; \hat{\rm Y})}{\sqrt{H(\rm Y) \times H(\hat{\rm Y})}}$,
where Y, $\hat{\rm Y}$, H and I are the true clusters, the predicted clusters, the entropy and the mutual information, respectively.

The second, known as the adjusted rand score or adjusted rand index (ARI), estimates a similarity between predicted clusters versus true clusters by considering all pairs of samples and counting pairs that are assigned correctly into the same or different clusters. The Rand index (RI)~\cite{randindex} is
defined as $RI = \frac{{S_1}+{S_2}} {\binom{M}{2}},$
where $M$ is the number of elements in the test set $U$, $S_1$ is the number of pairs of elements in $U$ that are in the same subset in the true clustering assignment and the predicted clustering assignment, and $S_2$ is the number of pairs of elements in $U$ that are in different subsets in the true clustering assignment and in the predicted clustering assignment.
The adjusted rand index is the corrected-for-chance version of RI. The adjusted rand index can yield negative values if the index is less than the expected index. The ARI is defined as $ARI = \frac{(RI - \textrm{Expected}_{RI})} {(\textrm{max}(RI) - \textrm{Expected}_{RI})}$.

\subsection{Dataset}
We use Gravity Spy Dataset 1.0 presented and discussed in detail in~\cite{datasetpaper}, which uses data from the Hanford and Livingston detectors during the first and second observing runs of advanced LIGO. An earlier version of this dataset is used in~\cite{bahaadini2017deep, zevin2017gravity}. This dataset has $21$ morphologically distinct glitch classes, and one catch-all `none of the above' class. We do not use this class in our experiment as this class is ill-defined in terms of its morphological features. From the $21$ distinct glitch classes, we randomly select $5$ of them as the ``unknown'' classes. The other $16$ classes are used for training DIRECT. It is important to emphasize that these two sets of classes are totally disjoint --  all algorithms are trained with the samples of known glitch classes and tested on unknown classes. 
\subsection{Baseline}
We compare the performance of our algorithm with the following baselines. 
The number of clusters is set equal to the number of unknown classes. Although in the real application of DIRECT we may not know the exact number of clusters beforehand, this exercise is meant to compare DIRECT against other representation algorithms and this assumption suits that purpose.
\begin{table}[h]
\centering
\caption{The performance comparison of DIRECT with the presented baselines. The best performances are bold for each evaluation measure.}
\label{tab:res}
\begin{tabular}{l|ll}
 Method & NMI & ARI \\ \hline
Original feature & 0.5131 & 0.1986 \\
PCA &0.5117  & 0.1938 \\
Deep Autoencoder & {{0.5451}} & {{0.3243}} \\
DIRECT (proposed model) & $\textbf{0.5978}$ & $\textbf{0.4550}$ 
\end{tabular}
\end{table}
\begin{itemize}
\item \textbf{Raw features:}
\begin{figure}[b]
\centering
\includegraphics[width=0.33\textwidth]{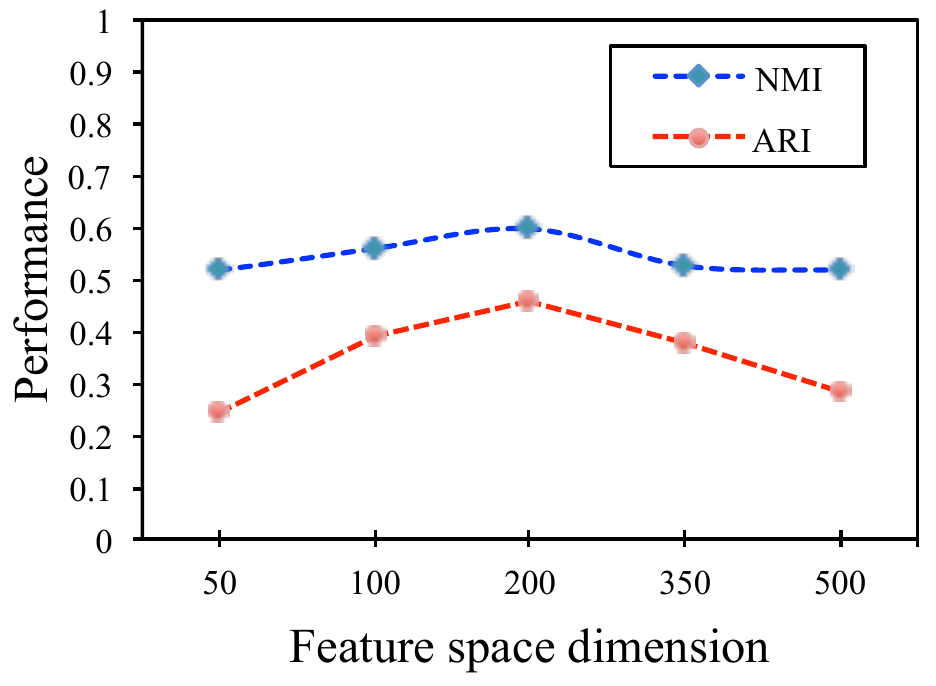}
\caption{The effect of size of feature space $k$ on DIRECT performance.}
\label{fig:size}
\end{figure}
\begin{figure*}[h!]
\begin{center}
\begin{subfigure}{.41\textwidth}
  \centering
\includegraphics[width=.98\linewidth]{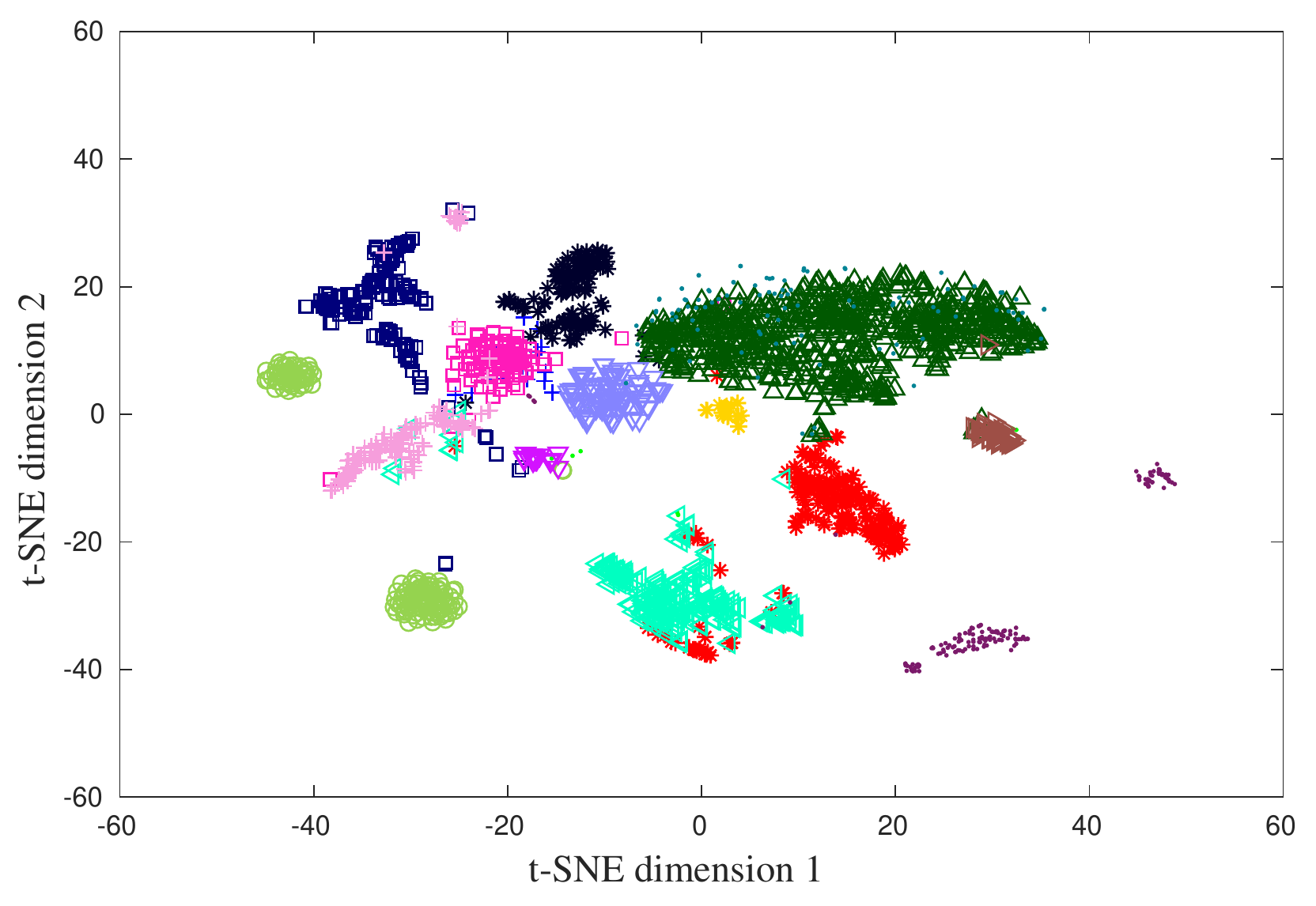}
\caption{Raw Feature}
  \label{fig:mnistS}
\end{subfigure}%
\begin{subfigure}{.41
\textwidth}
  \centering
  \includegraphics[width=.98\linewidth]{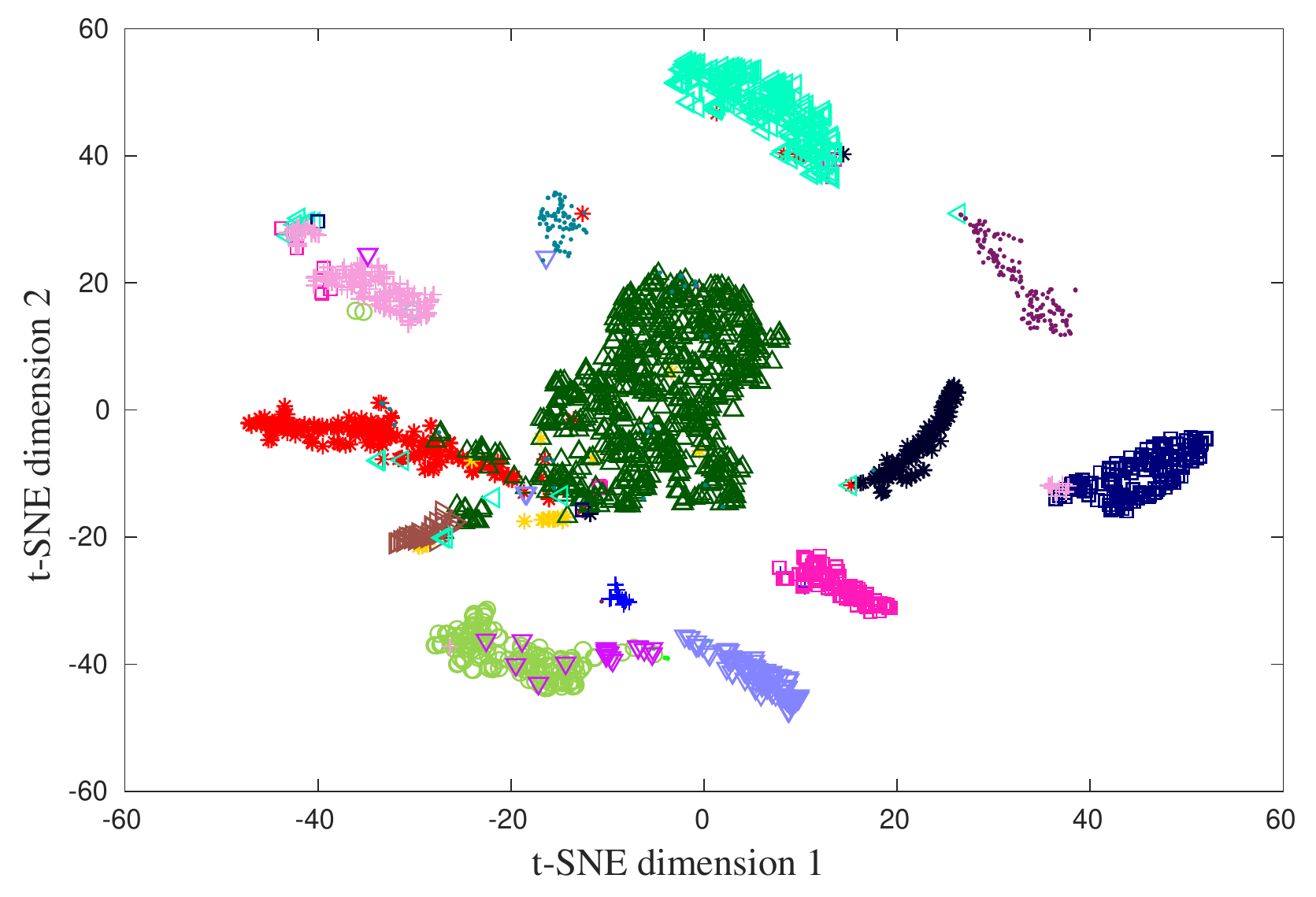}
  \caption{DIRECT}
  \label{fig:lfwS}	
\end{subfigure}
\end{center}
\caption{Visualization of various feature spaces. The right plot, obtained from the proposed model, is more discriminative and many of the scattered classes seen in the raw feature space (left plot) are consolidated to form a coherent class.}
\label{fig:tsne}
\end{figure*}
The first baseline performs the clustering task on the original feature space and evaluates how the efficiency of raw features for clustering and finding new classes. 
\item \textbf{Principal component analysis (PCA):}
Principal component analysis is one of the standard dimensional reduction techniques which converts the original dataset into a dataset with linearly uncorrelated variables, known as principal components. 
Here, we use PCA to find the directions with the most variance in the known classes samples.
We selected the first $50$ principal components, which capture approximately $98\%$ of the variance existing in the data. 
\item \textbf{Deep autoencoder:}
Autoencoder has been used for unsupervised representation learning. 
We can split the autoencoder into two parts: an encoder and a decoder. The encoder maps the input to an abstract, low-dimensional feature space, and the decoder maps the abstract feature space back to the original feature space.
We use the features extracted by the encoder as our deep autoencoder comparison.
\end{itemize}
\vspace{-1mm}
The performance of these methods are compared in Table~\ref{tab:res}. As we expected, the performance of PCA and raw data are very close. This is mainly because the principal components learned on the space of known classes are not necessarily generalizable for the unknown classes space. DIRECT gives the best result, showing the ability of this model in transferring labeled data in a separate domain for the target clustering task. Although deep autoencoder shows better performance than PCA and raw feature, it still cannot use the labeled data information in the way DIRECT can as it is an unsupervised representation learning technique.

\subsubsection{Deep learning setting}
For all our deep learning implementations, we use Python utilizing the Keras library~\cite{chollet2015}.

\vspace{+2.5mm}
\noindent
\textbf{DIRECT configuration:} \\
For learning $f_{\theta}$, we first use the convolutional layers with weights of pre-trained vgg16 network~\cite{simonyan2014very} and add two fully connected layers with sizes of $1024$ and $200$ using linear and ReLU activation functions, respectively. The fully connected layer with a linear activation function has a kernel $l_2$ regularizer of $10^{-4}$. The objective function is given in detail in Section~\ref{sec:obj}. The batch size is set to $20$ and the number of epochs is set to $50$ for DIRECT. 
\vspace{+2.5mm}

\noindent
\textbf{Deep Autoencoder configuration:}\\
For deep autoencoder, we use three fully connected layer with size of $300$, $200$, and $300$ with sigmoid activation function. The original glitch image sizes are down-sampled to $150\times150\times3$. The objective function of the deep autoencoder is binary cross entropy. The batch size is set to $20$ and the number of epochs is $50$.
\vspace{-1mm}
\subsection{Model Analysis}
\vspace{-1mm}
\subsubsection{Size of feature space}
We investigate the effect of feature space size~\cite{yang1997comparative} on the performance of DIRECT in Fig.~\ref{fig:size}. 
We see that increasing the size of $k$ enables the algorithm to better learn the unknown sample space through learning the known sample space, but after a certain point the size of $k$ becomes too large and DIRECT becomes prone to overfitting. We have determined that $200$ dimensions is the ideal value for $k$, as it provides the clustering algorithm enough information. After $200$ dimensions, it seems that the neural network is overfit on known classes and it may maintain noise or other irrelevant information that make the clustering less generalizable to unknown classes. 
\subsubsection{Visualizing feature space}
Using the t-distribution stochastic neighbor embedding (t-sne)~\cite{tsne} algorithm, the feature space obtained from DIRECT is visualized and compared with the original feature space in Fig.~\ref{fig:tsne}. 
Examining the DIRECT feature space, we observe that samples of certain classes which are scattered in the original feature space are more tightly clustered. As an example, the two segregated clusters of the class represented by light green circles in the raw feature space (left and bottom left of the raw feature plot) are merged together as a distinct class in the feature space obtained from DIRECT space (bottom left of the DIRECT feature plot). This merging of segregated clusters can also been seen with the class represented by the small purple circles (bottom-right of raw feature plot, center-right of DIRECT feature plot). In addition to merging segregated clusters, DIRECT generally tightens the feature-space clustering of classes, as can be seen by the class represented by dark blue squares (upper-left of raw feature plot, center-right of DIRECT plot). 
\vspace{-4mm}
\section{Conclusion}
\vspace{-1.9mm}
\label{sec:conc}
We present a deep discriminative representation for clustering of LIGO data. A embedding function is used to transfer knowledge from a set of known glitch classes to unknown glitch classes. The parameters of this nonlinear function are learned by utilizing a deep neural network. This function maps samples to a discriminative feature space where a clustering algorithm can work more efficiently. We compare our framework with three baselines, outperforming all of them.
\vspace{-1.5mm}
\section{Acknowledgement}
This work was supported in part by an NSF INSPIRE grant (award number IIS-1547880) and IDEAS Data Science Fellowship, supported by the National Science Foundation under grant DGE-1450006.
\bibliographystyle{IEEEbib}
\bibliography{strings,icip}

\end{document}